\documentclass{article}

\usepackage[final]{neurips_2021}

\usepackage[utf8]{inputenc} 
\usepackage[T1]{fontenc}    
\usepackage{hyperref}       
\usepackage{url}            
\usepackage{booktabs}       
\usepackage{amsfonts}       
\usepackage{nicefrac}       
\usepackage{microtype}      
\usepackage{xcolor}         
\usepackage{soul}

\title{From Convolutions towards Spikes: The Environmental Metric that the Community\\ currently Misses}

\author{%
  Aviral~Chharia\thanks{Authors claim equal contribution} \ , Shivu Chauhan$^{*}$
  , Rahul~Upadhyay, Vinay~Kumar \thanks{Corresponding Author: \texttt{vinay.kumar@thapar.edu}} \\
  Thapar Institute of Engineering and Technology, India\\
}

\begin{document}

\maketitle

\begin{abstract}
    Today, the AI community is obsessed with `state-of-the-art' scores ($80\%$ papers in NeurIPS) as the major performance metrics, due to which an important parameter, i.e., the environmental metric, remains unreported. Computational capabilities were a limiting factor a decade ago; however, in foreseeable future circumstances, the challenge will be to develop environment-friendly and power-efficient algorithms. The human brain, which has been optimizing itself for almost a million years, consumes the same amount of power as a typical laptop. Therefore, developing nature-inspired algorithms is one solution to it. In this study, we show that currently used ANNs are not what we find in nature, and why, although having lower performance, spiking neural networks, which mirror the mammalian visual cortex, have attracted much interest. We further highlight the hardware gaps restricting the researchers from using spike-based computation for developing neuromorphic energy-efficient microchips on a large scale. Using neuromorphic processors instead of traditional GPUs might be more environment friendly and efficient. These processors will turn SNNs into an ideal solution for the problem. This paper presents in-depth attention highlighting the current gaps, the lack of comparative research, while proposing new research directions at the intersection of two fields- neuroscience and deep learning. Further, we define a new evaluation metric `$\mathbf{NATURE}$' for reporting the carbon footprint of AI models. 
\end{abstract}

\underline{\textbf{The Environmental perspective that models cannot overlook.}} Training an NLP model could generate up to $626,155$ pounds of CO$_{2}$ emissions—roughly equal to the total lifetime carbon footprint of $5$ cars [1]. It is evident from the recently developed OpenAI's GPT-$3$ [2] (with $175$ billion parameters) that required several thousand petaflop days ($355$ years on standard GPU) to train. Moreover, with such ever-increasing model sizes, model performance diminishes relatively faster. ResNeXt [3] required 35\% higher computational resources compared to ResNet [4] to achieve a mere 0.5\% improvement in accuracy. Such diminishing model returns is also seen in GPT-3 compared to GPT-2 [5]. Currently, more than 80\% of papers in NeurIPS and 75\% in CVPR target accuracy as the major improvement, whereas only a very small group of papers (<20\% in CVPR) argue for a new efficiency result [6]. This clearly establishes the AI community's present obsession with obtaining SOTA scores on performance benchmarks. This is the key reason why the community prioritizes performance metrics like accuracy over efficiency metrics like speed and model size. 

\underline{\textbf{Why Current Models are not what Hubel \& Wiesel studied.}} Hubel and Wiesel [7] demonstrated that neurons in a cat's primary visual cortex are tuned to simple features. However, their studies were based on unsupervised recognition models, whereas most models designed today are supervision-based. This is primarily the reason why current ANNs are not directly inspired by nature, and when compared to human brain (that consumes just $20$W power),  require substantial energy resulting in high environmental costs. R\&D of new model architectures multiply this energy requirement by $1000\times$ due to retraining of model hyperparameters during experiments. Tweaking network design and using fewer neurons/layers could hardly solve the computational problem effectively.

\underline{\textbf{A field comparatively less Researched.}} Neuromorphic computing approaches the problem of low-power model design quite differently and has two characteristics: (i) developing brain-inspired nets, (ii) designing efficient hardware for such algorithms. Neurons in SNNs, like those in nature, interact with each other via isolated, discrete electrical impulses (spikes) rather than continuous signals, and function in continuous rather than in discrete-time. Instead of modeling the dynamics of the model on standard von-Neumann computers, neuromorphic hardware is particularly intended to operate such networks with very little power overhead, using electronic circuits that correctly recreate the dynamics of the model in real-time. Due to difficulty in communicating weights between network nodes, current backpropagation based ANNs are challenging to build on non-von-Neumann neuromorphic architecture. SNNs, on the other hand, are trained without supervision (except output layer) using spike-timing-dependent plasticity. Here, the biological synapse's synaptic plasticity has access to the activity of the two neurons it links, but not to the activity of other neurons to which it is not physically linked. Due to this, the plasticity of the brain is local. 

\underline{\textbf{From Convolutions to Spikes.}} 
In a SNN with `$\mathbf{N}$' pre-synaptic neurons, the post synaptic potential of the $i^{th}$ neuron is computed as the product of input spike signal $s_i(t)=0$ or $1$ with the synapse weight $w_i$. At any time $t$, if $\mathcal{V}\mathbf{(t)}$ i.e., the neuron membrane potential, surpasses the set threshold, the output neuron spikes with $\sum_{i=1}^{\mathbf{N}} w_k s_k (t)$. The weights can be updated according to an unsupervised learning STDP rule that results in output spiking if a fixed pattern $S_{fixed}$ is present: ($w_i\leftarrow w_i+\Delta w_i, \Delta w_i = -a^- w_i(1-w_i)$, if $t_{out}-t_i<0$ else $= +a^+ w_i(1-w_i)$, if $t_{out}-t_i\geq0$). Contrary to SNNs, CNNs are based on convolution operations with a much more accumulative membrane potential given by, $\mathcal{V}_{\mathbf{m}}(u, v, t, 1)=\sum_{\tau=0}^{\mathbf{t}} \big( \sum_{j=-2}^{2} \sum_{i=-2}^{2}s_{in}(u+i, v+j, \tau)W_{C1}(i, j, 1) \big)$. The first convolutional layer neuron fires at time `$t$’ if $\mathcal{V}_{\mathbf{m}}^{(1)}(u, v, t)\geq\gamma_{C1}$. It is evident that CNNs have far higher computational complexity compared to SNNs. Thus, SNNs are highly energy-efficient, computationally less intensive; and, designing models based on them are comparatively more environment friendly. DL-simulated SNNs are recently utilized for image classification tasks. One of the major advantages of converting CNNs to spiking-CNNs is that it allows to use SNNs' sparse computing and perform similar computation with less energy. With this, SNNs can achieve $\times 38.7$ times better energy efficiency than ANNs without any significant loss in accuracy [9]. 

\underline{\textbf{A Novel Evaluation Metric.}} We suggest that the community must report the model training duration, hyper-parameter sensitivity, environmental costs associated as an equally important metric while proposing new models. We propose a novel evaluation metric `$\mathbf{NATURE}$'\footnote{Here, $\mathbf{N}_{exp}$ = Number of experiments, $\mathbf{A}$ = Constant to account for energy loss, $\mathbf{T}$ = time in sec, $\mathbf{\mathcal{U}}_{datacenter\ utility}$ = Data center energy utility, $\mathcal{R}_{Regional\ grid}$ = Energy consumption from Regional Grid, $\mathcal{E}_{hardware}$ = Hardware energy consumption, epochs = Number of training epochs} \begin{center}
    $\mathbf{NATURE} = \mathbf{N}_{exp} \times [\mathbf{A} + \mathbf{T} \times (\mathbf{\mathcal{U}}_{datacenter\ utility}+\mathcal{R}_{Regional\ grid} + \mathcal{E}_{hardware}) \times$ epochs]
\end{center}\underline{\textbf{Neuromorphic Computing- A tale of Hardware.}} Implementing SNNs on neuromorphic hardware instead of traditional GPUs is found to be more environmental friendly since it runs asynchronously using spikes. The conventional von Neumann computing architecture separates processing from the memory unit. CMOS-based chips (by IBM and Intel) for SNN computations are limited to large-scale modular computing units with increasing neurons and synapses. The current study reiterates that significant advances are required to increase onboard neurons and synapses to mimic complex operations of the brain like cognitive processing, sophisticated motor control, learning, and abstraction. Moreover, CMOS-based systems are limited by their room-scale size. Thus, we need to look into Memristor-based alternatives due to their synaptic-like behavior. Memristor, i.e., `Resistor+memory,' is a building block of synaptic devices that requires high integration density, low latency, low power, and nonvolatile memory capable of mimicking brain. These are imperative for implementation of synaptic learning used in neuromorphic computing. These observations support Memristors as promising candidates for future synaptic learning applications [8].

\underline{\textbf{The Hardware Gap and Proposed Opportunities.}} New synaptic memory devices exhibit issues that make older design principles obsolete. The resistance variation of ReRAM (Resistive RAM) and PRAM (Phase Change Memory RAM) introduces errors over time, thus a trade-off between energy consumption and induced error has to be maintained. A high error rate may be suitable in some hardware systems as long as the system's core functionality is met, in our case, lesser energy consumption to protect environment. Hybrid systems of memristor layers embedded on CMOS substrate is solution to neuromorphic hardware engineering [8]. This will enable computing and learning processes by combining memristors on spiking processors to fire neurons in silicon chips after attaining a specific threshold value [10]. PCB design techniques [11] can assemble chips with memristors, substantially scaling up number of neurons and synapses in a neural system. However, there would be limitations relating to memristor layers density and their onboard program-ability.

\underline{\textbf{Conclusion and Future Research Directions.}} Along with developing new SOTA approaches, emphasis should be placed on building more energy-efficient algorithm. The current study emphasizes and draws the community's attention to this endeavor. The study further highlights the lack of comparative research in using SNN-based approaches for more computation-efficient algorithms. Moreover, the existing gaps between accessible RAM (Hybrid-CMOS-Memristor Architecture) choices and the optimum hardware required for Neuromorphic engineering were explored in this article. We support the view that hybrid Memristor-CMOS-based PCB designs may significantly increase the number of neurons and synapses, and more accurately simulate the brain. Moreover, we suggest that the community must report environmental costs associated as an equally important metric while proposing new models. For the same, we proposed a novel evaluation metric (`$\mathbf{NATURE}$'). Further, focus must be on hardware and methods that are computationally efficient. More efficient computing will facilitate a shift from data centres to edge devices, allowing AI to reach a larger audience while minimising data leakage, lowering transmission costs, and boosting privacy and inference speeds.

\section*{References}

\medskip

{
\small

[1] Strubell, E., Ganesh, A., \& McCallum, A. (2019). Energy and policy considerations for deep learning in NLP. Retrieved from http://arxiv.org/abs/1906.02243

[2] Tom B. Brown, et al. (2020). Language Models are Few-Shot Learners. http://arxiv.org/abs/2005.14165

[3] Xie, S., Girshick, R., Dollár, P., Tu, Z., \& He, K. (2017). Aggregated residual transformations for deep neural networks. In Proceedings of the IEEE conference on computer vision and pattern recognition (pp. 1492-1500).

[4] He, K., Zhang, X., Ren, S., \& Sun, J. (2016). Deep residual learning for image recognition. In Proceedings of the IEEE conference on computer vision and pattern recognition (pp. 770-778).

[5] Radford, A., Wu, J., Child, R., Luan, D., Amodei, D., \& Sutskever, I. (2019). Language models are unsupervised multitask learners. OpenAI blog, 1(8), 9.

[6] Schwartz, R., Dodge, J., Smith, N. A., \& Etzioni, O. (2020). Green ai. Communications of the ACM, 63(12), 54-63.

[7] Wiesel, T. N., \& Hubel, D. H. (1963). Effects of visual deprivation on morphology and physiology of cells in the cat's lateral geniculate body. Journal of neurophysiology, 26(6), 978-993.

[8] Kundu, S., Datta, G., Pedram, M., \& Beerel, P. A. (2021). Spike-thrift: Towards energy-efficient deep spiking neural networks by limiting spiking activity via attention-guided compression. In Proceedings of the IEEE/CVF Winter Conference on Applications of Computer Vision (pp. 3953-3962).

[9] Park, J. (2020). Neuromorphic Computing Using Emerging Synaptic Devices: A Retrospective Summary and an Outlook. Electronics, 9(9), 1414.

[10] Camuñas-Mesa, L. A., Linares-Barranco, B., \& Serrano-Gotarredona, T. (2019). Neuromorphic spiking neural networks and their memristor-CMOS hardware implementations. Materials, 12(17), 2745.

[11] Zamarreño-Ramos, C., Linares-Barranco, A., Serrano-Gotarredona, T., \& Linares-Barranco, B. (2012). Multicasting mesh AER: A scalable assembly approach for reconfigurable neuromorphic structured AER systems. Application to ConvNets. IEEE transactions on biomedical circuits and systems, 7(1), 82-102.

}

\end{document}